\documentclass[runningheads]{llncs}

\usepackage{graphicx}
\usepackage[hidelinks]{hyperref}
\usepackage{soul}
\usepackage{xcolor}

\begin{document}

\title{An Interactive Interface for Novel Class Discovery in Tabular Data}
% Alternative: A Visual Interface for Novel Class Discovery in Tabular data
\author{Colin Troisemaine\inst{1, 2} \and
Joachim Flocon-Cholet\inst{1} \and
Stéphane Gosselin\inst{1} \and
Alexandre Reiffers-Masson\inst{2} \and
Sandrine Vaton\inst{2} \and
Vincent Lemaire\inst{1}}

\authorrunning{C. Troisemaine et al.}

\institute{Orange Innovation, Lannion, France \and
Department of Computer Science, IMT Atlantique, Brest, France
\email{colin.troisemaine@orange.com}}

\maketitle

\begin{abstract}
Novel Class Discovery (NCD) is the problem of trying to discover novel classes in an unlabeled set, given a labeled set of different but related classes.
% While numerous methods have been proposed for image data, only one paper has tackled the case of tabular data, despite being among the most widely used type of data in practical applications.
The majority of NCD methods proposed so far only deal with image data, despite tabular data being among the most widely used type of data in practical applications.
To interpret the results of clustering or NCD algorithms, data scientists need to understand the domain- and application-specific attributes of tabular data.
This task is difficult and can often only be performed by a domain expert.
Therefore, this interface allows a domain expert to easily run state-of-the-art algorithms for NCD in tabular data.
With minimal knowledge in data science, interpretable results can be generated.

\end{abstract}

\section{Introduction}
\label{sec:intro}

Novel Class Discovery (NCD) \cite{hsu2018learning,tr2023introduction} is a new and growing field, where we are given during training a labeled set of known classes and an unlabeled set of different classes that must be discovered.
In recent years, many methods have been proposed in the context of computer vision \cite{chi2022meta,autonovel2021,han2019learning}.
% An intuitive example of application would be: a labeled set of images of some dog breeds should help guide a system identify other dog breeds in an unlabeled set.

Tabular data refers to data arranged in a table, where each row is an observation and each column is an attribute.
It is one of the most common types of data in practical applications such as medical diagnosis, customer churn prediction, cybersecurity, and credit risk assessment. \cite{shwartz2022tabular}.
An intuitive example of application of NCD in tabular data would be customer churn prediction: by using a dataset that includes the reasons why customers stopped using a product, we can more accurately identify other causes of churn in an unlabeled set where the reasons have not yet been identified.

While in practice, tabular data is one of the most prevalent data types in the real world, to the best of our knowledge, only one paper has attempted to solve NCD specifically for tabular data \cite{tabularncd}.
This is partly due to the heterogeneous nature of tabular data, and its lack of spatial and semantic structure, which makes it difficult to apply some computer vision techniques such as data augmentation or Self-Supervised Learning \cite{deeptabularsurvey}.
Furthermore, tabular data contains attributes that are specific to each domain.
This means that analyzing and understanding the results of NCD or clustering algorithms can be challenging for a data scientist who is not necessarily familiar with the attributes of the dataset.
On the other hand, the domain expert does not necessarily have the knowledge required to write code and run NCD or clustering algorithms.

In an ideal scenario, the domain expert would be included in the training loop to interpret the results produced by the data scientist.
But for practical reasons, it can be difficult to dedicate two people to this task, as having a data scientist run an algorithm, present the results to the expert, and update the parameters based on the expert's feedback can be a slow and tedious process.

Hence, the goal of the interface proposed here is to allow a domain expert to visualize his data and run NCD or clustering algorithms without having to write code, as in visual data mining \cite{soukup2002visual}.
Given a pre-processed dataset, a user can employ this interface to (i) get a first idea of the separability of the data with T-SNE, (ii) select which features and classes to use, and which classes are considered unknown (iii) parameterize and execute NCD and clustering algorithms and (iv) train decision trees to generate rules and interpret the classes or clusters.
Based on theses results, an expert can remove features or classes that have too much influence on the results, re-train a clustering model and re-generate rules.
This process can be very tedious through code, but it can be done in only a few clicks with this interface (which even a data scientist could benefit from).

Currently, this interface implements TabularNCD~\cite{tabularncd}, the state-of-the-art for NCD in the context of tabular data.
Other clustering methods are implemented: spectral clustering, $k$-means and a simple baseline method to solve NCD.
This baseline trains a classification neural network on the labeled data, and then projects the unlabeled data in its last layer before clustering it with $k$-means.

As expressed before, this interface cannot replace the domain expert.
It only allows him to explore his dataset using machine learning tools without writing code.
This interface is also upgradeable, as new NCD or clustering algorithms can be quickly implemented.
% Note that while there exists some automated clustering software, none are adapted to solve the specific NCD problem.
The application is open source and can be installed locally using the code at \url{https://github.com/ColinTr/InteractiveClustering}.
The video of the demonstration is available at \url{www.youtube.com/watch?v=W7ru8NHPj-8}.

\section{Interface description}

\begin{figure}[htbp]
    \centering
    \includegraphics[width=\textwidth]{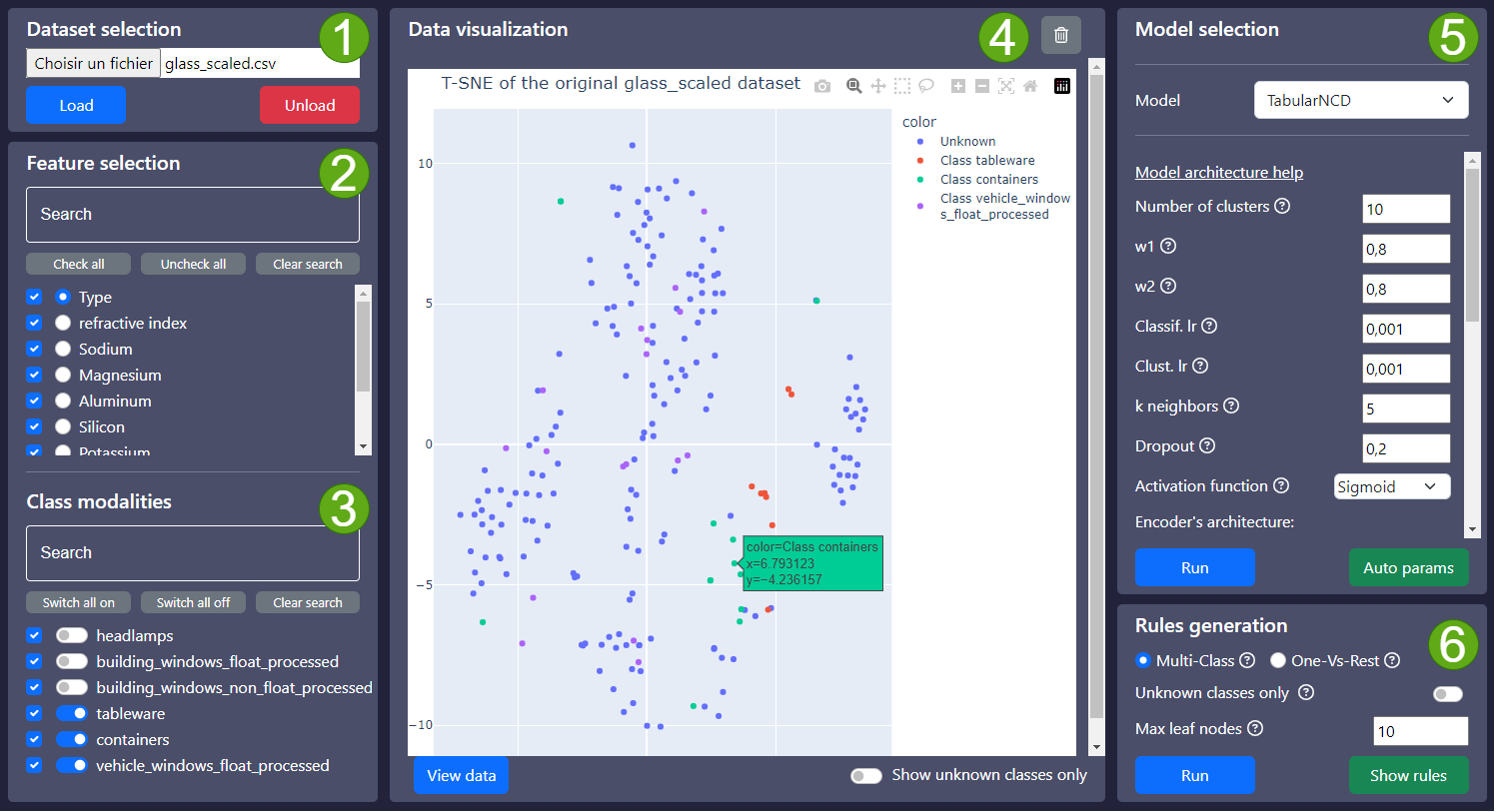}
    \caption{The interface for interactive clustering and Novel Class Discovery.}
    \label{fig:interface}
\end{figure}

As shown in Figure~\ref{fig:interface}, the interface is composed of 6 different panels that we will describe in this section.
For reference, the interface was made in JavaScript with React 18.2.0, and the Python code is executed by a Flask 2.2.2 backend server.

After selecting and loading a dataset with panel (1), the user can select in panel (2) which features to use in the dataset, and indicate which is the class feature.
Panel (3) lists the modalities of the class feature picked earlier.
Here, the user can choose to remove some classes from the dataset by unchecking them and select which classes are considered as known or unknown.
In a use-case with a real dataset including both labeled and unlabeled data, a group of observations could be labeled as ``unknown'', which can thus be selected in this panel.

With panel (4), the data can be visualized in 2 dimensions by running a T-SNE.
The user also has the option to view only the unknown classes for easier readability.
Clicking on a point displays all its attributes.
Note that in an effort of optimization and better responsiveness, if a data plot is requested and has the same coordinates as a previous request, the T-SNE is re-used and only the coloring of the points is updated.

The NCD and clustering models can be selected and configured in panel (5).
Currently, 4 models are available:
TabularNCD~\cite{tabularncd} is a NCD method that pre-trains a simple encoder of dense layers with the VIME \cite{vime} self-supervised learning method.
It adopts an architecture with two ``heads'': one to classify the known classes and introduce relevant high-level features in the latent space of the encoder, and another classifier for the unlabeled data trained with pseudo-labels defined without supervision in the latent space.
Next is $k$-means, which was implemented for its simplicity and wide adoption in the community.
It has the advantage of having a single parameter (the number of clusters).
Spectral clustering is also available.
It is known for its good results and its ability to discover new patterns across a wide variety of datasets \cite{NIPS2001_801272ee}.
And finally the baseline method described in Section~\ref{sec:intro} can be selected.
Both TabularNCD and the baseline rely on an architecture composed of a combination of dense layers, dropout and activation functions which can all be modified through the interface (even the sizes and number of hidden layers).

Starting the training of TabularNCD or the baseline will produce a pop-up that displays the current progress of the training and the estimated time to completion.
It is also possible to visualize a T-SNE of the latent space of these models, instead of visualizing the original features of the data.

\begin{figure}[htbp]
    \centering
    \includegraphics[width=\textwidth]{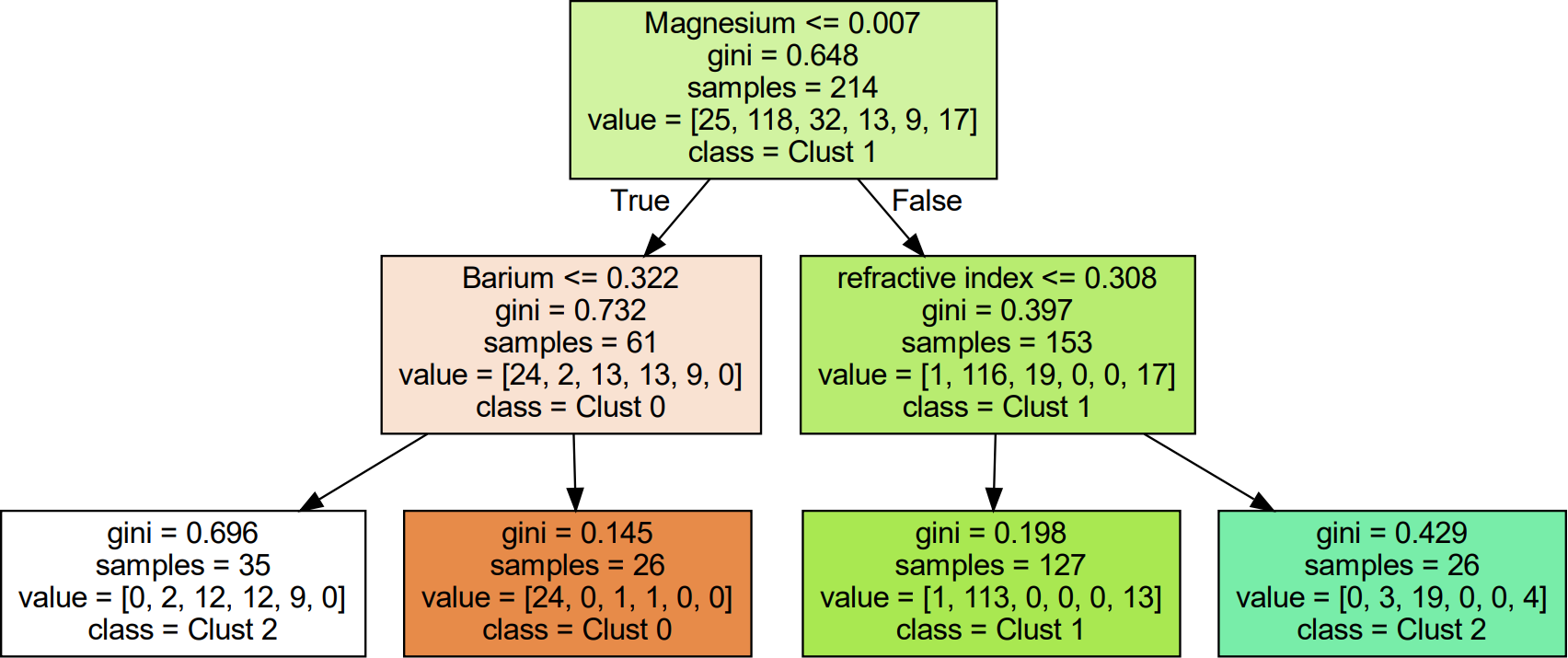}
    \caption{Example of rules that describe the classes of the \textit{glass identification} dataset.}
    \label{fig:rules}
\end{figure}

Finally, in panel (6), the user can get an interpretable description of the results by training a decision tree to classify the known classes and the discovered clusters.
Figure~\ref{fig:rules} is an example of rules in a decision tree obtained for the \textit{glass} dataset.
Each box represents a node/leaf of the tree and displays the rule and the majority class.
The tree can be multi-class and will give an overview of the relations between all the classes and clusters, but it can be hard to comprehend because of its complexity.
For this reason, we can instead use a \textit{one-versus-rest} approach, where for each class or cluster, a decision tree has to predict the class or cluster against all the others.
As each individual tree solves a problem of lower complexity, they are shorter compared to the multi-class case and are more easily interpretable.

\section{Conclusion}

This demo paper introduces an interactive interface for the problem of Novel Class Discovery in tabular data.
This interface is mainly targeted to domain experts and data scientists.
The user can quickly visualize the data and generate clusters of novel classes along with interpretable decision trees to describe them.
Furthermore, the user can easily identify both features and classes to remove from the training process and start a new clustering with different parameters.

In the future, this interface could be improved by adding a function to estimate the number of clusters (i.e. the number of novel classes).
New NCD and clustering methods can also be easily integrated.
Giving the user the ability to merge or split some clusters and update the decision tree's rules accordingly could also be an interesting addition.

\bibliographystyle{splncs04}
\bibliography{bibliography}

\end{document}